\documentclass[letterpaper]{article} 
\usepackage[]{aaai24}  
\usepackage{times}  
\usepackage{helvet}  
\usepackage{courier}  
\usepackage[hyphens]{url}  
\usepackage{graphicx} 
\usepackage{natbib}  
\usepackage{caption} 
\usepackage{caption}
\usepackage{dirtytalk}
\usepackage{graphicx}
\usepackage{textcomp}
\usepackage{subfig}
\usepackage[detect-all]{siunitx}

\usepackage{aligned-overset}
\usepackage{array, multirow}
\usepackage{float}
\usepackage{tabularx}

\usepackage{amsmath}
\usepackage{amssymb}
\usepackage{mathtools}

\usepackage{newfloat}
\usepackage{listings}
\DeclareCaptionStyle{ruled}{labelfont=normalfont,labelsep=colon,strut=off} 
\floatstyle{ruled}
\newfloat{listing}{tb}{lst}{}
\floatname{listing}{Listing}
\title{Deep Learning-based Group Causal Inference in Multivariate Time-series}
\author{
    Wasim Ahmad,
    Maha Shadaydeh, 
    Joachim Denzler 
}
\affiliations{
    Computer Vision Group, Faculty of Mathematics and Computer Science, \\
    Friedrich Schiller University Jena, Germany\\
    \{wasim.ahmad, maha.shadaydeh, joachim.denzler\}@uni-jena.de

}

\usepackage{bibentry}

\begin{document}
\maketitle

\begin{abstract}

Causal inference in a nonlinear system of multivariate time series is instrumental in disentangling the intricate web of relationships among variables, enabling us to make more accurate predictions and gain deeper insights into real-world complex systems. Causality methods typically identify the causal structure of a multivariate system by considering the cause-effect relationship of each pair of variables while ignoring the collective effect of a group of variables or interactions involving more than two-time series variables. In this work, we test model invariance by group-level interventions on the trained deep networks to infer causal direction in groups of variables, such as climate and ecosystem, brain networks, etc. Extensive testing with synthetic and real-world time series data shows a significant improvement of our method over other applied group causality methods and provides us insights into real-world time series. The code for our method can be found at: https://github.com/wasimahmadpk/gCause.
\end{abstract}
\section{Introduction}
\label{intro} 
\begin{figure}[t]
\centering
\includegraphics[width=0.33\textwidth]{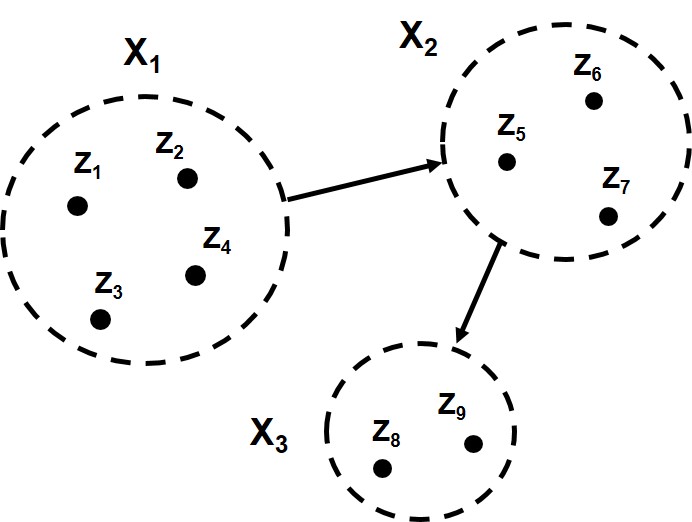}%
\caption{Interactions among groups of multivariate time series data of various dimensions.} 
\label{fig:gcause}
\end{figure}

Group-based causal inference investigates causal relationships within specific groups of individuals, entities, or units. It is particularly relevant when studying complex systems with interconnected components such as in climate \cite{molotoks2020comparing}  and brain networks \cite{faes2022new}, allowing researchers to investigate how variables within distinct groups contribute to observed outcomes or behaviors. In this paper, we explore the fundamental concept behind group-based causal inference in multivariate time series and its growing significance in diverse disciplines. To this end, we present our approach to testing the causal link in groups of time series, see Figure \ref{fig:gcause}, which provides a comprehensive understanding of how these groups interact. The proposed method allows the testing of bi-directional causal links. Our method builds on our previous work \cite{ahmad2022causal}, which exploits the model invariance property through Knockoffs \cite{barber2015controlling, barber2019knockoff, barber2020robust} interventions for pairs of variables in deep networks for causal estimation. However, here we emphasises the identification of causal interaction in groups of variables and is the first group causality method to the best of our knowledge which utilize deep learning to learn complex nonlinear relation.  The model invariance property refers to the invariant behavior of the model in different settings when its causal predictors are observed \cite{peters2016causal}. We use DeepAR \cite{salinas2020deepar} to model multivariate time series which has the potential to learn complex interactions in nonlinear data. Since deep networks cannot handle missing variables or adjust to out-of-distribution data, we use Knockoff variables for interventions on the trained deep networks. Knockoffs are in-distribution, uncorrelated copies of the original data with similar covariance structure.

To demonstrate the robustness and practical utility of our approach, we evaluate its performance on synthetically generated time series and real-world datasets. We compare our method with other group causality methods, i.e., Vanilla-PC \cite{janzing2009telling}, Trace method \cite{zscheischler2012testing} and 2GVecCI \cite{wahl2023vector}. As a real-world application, we mainly focus on the climate system and assessing the direction of connections in brain networks. This involves considering the internal processes over a period of time involving interactions within and between different climate subsystems, i.e., ocean, land, ecosystem, atmosphere, etc. \cite{latif2009nino}. For example, how climate and ecosystem interact \cite{malhi2020climate, sefidmazgi2020causality, korell2020we}, or how major climate phenomena such as the El Niño Southern Oscillation (ENSO) affect different global regions \cite{nowack2020causal}. These highly complex interactions provide the basis for understanding the profound impacts of environmental change. We conducted experiments on the FLUXNET data \cite{pastorello2020fluxnet2015}, which contains measurements from different sites that capture climate-ecosystem dynamics, where we aim to uncover the causal pathways between climate and ecosystem. We have also tested our method on a climate science dataset (ENSO 3.4), where we aim to estimate the impact of temperature patterns in the tropical Pacific on British Columbia. Moreover, we apply our method to simulated fMRI time series to identify connections in the brain network and indicate their directionality.
\section{Related Work}
\label{related} 
There are many methods for time series causal inference with applications in various fields that rely on certain assumptions that limit their applicability \cite{yao2020survey}. These approaches treat each system variable separately and estimate its causal relation to other variables, overlooking the group or collective influence, e.g., causal relation in brain regions \cite{siddiqi2022causal}, ecosystem and climate subsystems, etc.  The work of \cite{besserve2018group} presents a group theoretical framework to assess the relationship between cause and mechanism using group transformations. Vanilla-PC \cite{janzing2009telling} estimates the direction of a causal link in two groups of time series; however, the method suffers from high dimensionality in terms of accuracy and computation time because it runs a higher number of conditional independence tests. The Trace method is introduced by \cite{zscheischler2012testing} to infer causal direction in two linearly interacting groups which works faster, however, its performance degrades in case of high nonlinear relationships in groups. The authors of \cite{wahl2023vector} present 2GVecCI for uni-directional causality between two groups where they combine the constraint-based approach for causal discovery with sparsity measures of the internal causal structure of the groups. The 2GVecCI method is based on two principles for causal relations between groups of time series. First, conditioning on the cause group does not lead to new conditional dependencies within the effect group. Second, conditioning on the effect group does not delete conditional dependencies within the cause group. 

The authors of \cite{faes2022new} assess higher-order interactions in networks of processes through time and frequency domain analysis. The work of \cite{sato2010analyzing} uses Granger causality for a set of time series to analyze the connectivity between brain regions, assuming that sets are linearly related to each other while real-world data are highly nonlinear. Approaches that depend on the non-Gaussian nature of noise for drawing conclusions about causal direction, for example, LinGaM \cite{shimizu2006linear}, become less effective due to the influence of averaging caused by the central limit theorem, which tends to make the averaged noise distribution more Gaussian. Furthermore, methods for group causality aim to transform data by aggregation or dimensionality reduction, where a significant amount of information may be lost and, more importantly, the existing causal structure \cite{spirtes2000causation} is disrupted. However, our work utilizing deep networks has the potential to learn complex relationships in high-dimensional data.
\begin{figure*}[t]
\centering
\includegraphics[width=0.85\textwidth]{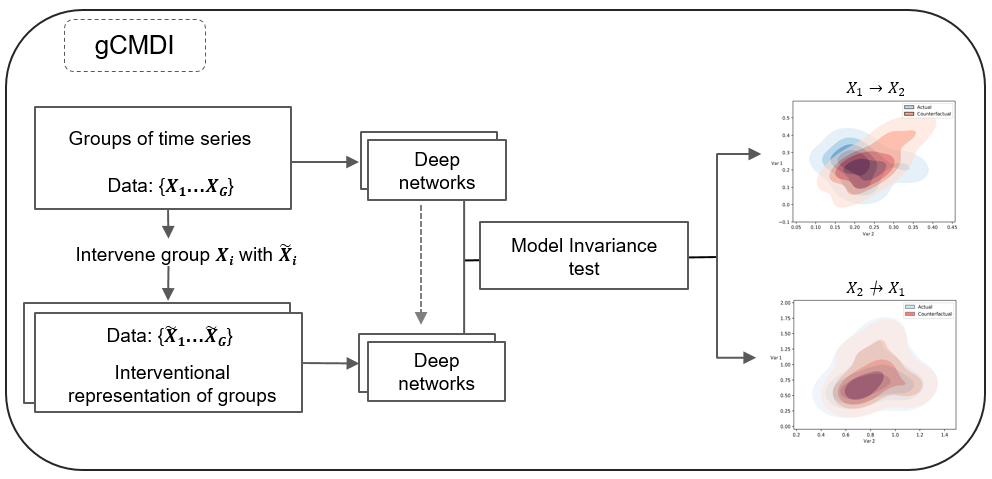}%
\caption{Schematic diagram of group causal discovery method where we model the complex relation among all variables in the system with deep learning approximator. For group-level causal inference, we implement model invariance testing by applying intervention to a group of interest and estimate its influence on the target group.} 
\label{fig:schematic}
\end{figure*}
For brain data, the detection of networks and their directions using fMRI time series is of high interest. The work of \cite{smith2011network} presents various correlation and causality-based method to identify the association between brain network nodes and assign a direction to them. They generate realistic simulated fMRI data for a wide range of underlying networks and experimental protocols to compare different connectivity estimation approaches. 
\begin{figure*}
\centering
\includegraphics[width=\textwidth]{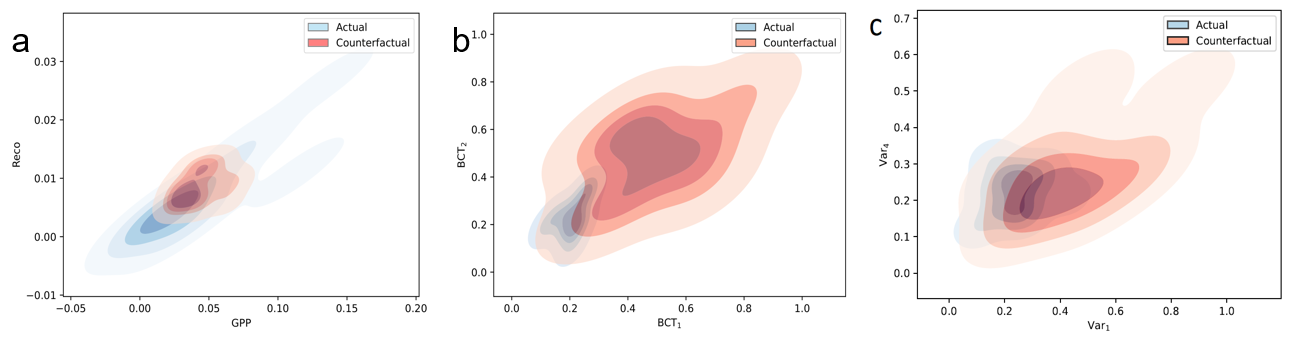}%
\caption{Distribution shift of  \textbf{a}.  ecosystem group $G_{\text{E}} = \{GPP, R_{\text{eco}}\}$ in response to intervention on climate group  $G_{\text{C}}= \{T, R_g\} $.  \textbf{b}. British Columbia temperature in response to intervention on ENSO temperature data. \textbf{c}. Brain network $N_2$ in fMRI time series after group-level intervention on network $N_1$.} 
\label{fig:jointdistshift}
\end{figure*}
\section{Method}
\label{methods}

In this section, we present the formulation of our method for group causal discovery assessing the causal relationships within sets of $N$-variate system of time series denoted as $Z^N = \{Z_1, \dots, Z_N\}$. These variables are organized into $G$ groups or subsets of variables, represented as $X^G = \{X_1, \dots, X_G\}$, which can be considered as indicative of the behavior of a network comprising $G$ distinct dynamic subsystems. Referring to the real-world applications, the groups $X_i, i=1, \dots, G$ may represent subsets of the climate system, brain networks, etc., where each group process $Z_{t, t=1, \dots, n}$, may represent the intra-atmospheric interaction or neural activity of a brain region. Each of these groups $X_i: Z \subset Z^N$ has a certain dimension $G_i$, such that the total number of variables $N$ can be expressed as $N = \sum_{i=1}^G G_i$. The method allows the assessment of uni- and bi-directional causal links in any number of groups. For two causally related groups $X_i, X_j, i, j=1, \dots, G$ where $i \neq j$, the possible links can be $X_i \overset{\text{causes}}{\longrightarrow} X_j$, or $X_j \overset{\text{causes}}{\longrightarrow} X_i$ or $X_i \overset{\text{causes}}{\longleftrightarrow} X_j$. 

\paragraph{Assumptions} 
We infer causal relationships in groups of nonlinear multivariate time series by making the following assumptions:
\begin{itemize}
\item[-] Stationarity: The variables $Z^N$ represent a stochastic stationary process. 
\item[-] Causal Sufficiency: The set of observed variables $Z^N = \{Z_1, \dots, Z_N\}$ contain all common causes in $Z^N$, i.e., no hidden confounders. 
\item[-] Model Invariance: The causal structure in the stochastic processes $Z^N$ remains consistent across different interventional enviroments.
\end{itemize}
The presented method for group causality in time series builds on our previous work \cite{ahmad2022causal} where we model the complex interactions in $N$-variate time series by training deep networks and apply model invariance testing through group-level interventions for inferring causal direction in groups. A model is invariant if, in the presence of its causal predictors, the distribution of its output residuals does not change across interventional environments \cite{peters2016causal}. The schematic diagram for our method is shown in Figure \ref{fig:schematic}, where we apply invariance testing to deep networks trained on sets of time series in order to infer causality in groups. Through deep autoregressive models \cite{salinas2020deepar} which takes $Z^N = \{Z_1, \dots, Z_N\}$ as input, we model the conditional distribution $P(Z_{i, t_0:T} | Z_{i, 1:t_0 - 1}, Z_{-i, 1:t_0 - 1}), i=1, \dots, N$ of the future of all system variables given their past values. Here $Z_{i}$ represents node of the target group and $Z_{-i}$ refers to the nodes in the rest of the groups in $X^G$. It utilizes recurrent neural networks (RNNs) to generate probabilistic output in terms of $\mu$ and $\sigma^2$ at each time point where the mean represents the central estimate, and the variance represents the uncertainty of the system. For all variables, we obtain distribution $R$ of their residuals $e^n=e_1, \dots, e_n$ for $n$ forecast windows where $e = \frac{1}{T}\sum_{t=1}^T |\frac{Z_t-Z_{pred}}{Z_t}|$ and $T$ represents forecast horizon. Pertaining to the invariance property, the model response $Z_i$, which is the target group of variables, does not change in different settings as long as we don't perturb the cause group.

\paragraph{Knockoff Intervention} 

The effectiveness of our previously proposed Knockoff intervention \cite{barber2015controlling, barber2019knockoff, barber2020robust} compared to other intervention methods is demonstrated on node-to-node causal relationship in multivariate time series \cite{ahmad2021causal, ahmad2022causal}. Let's say $X_j$ represent the group of output variables of the trained deep network, and $X_i$ denotes the input group where $i\neq j$. The causal effect of group $X_i$ on the network output group $X_j$ under intervention is denoted as: 

\begin{equation}
E(X_j|do(X_i=\tilde{X_i})    
\end{equation}


\begin{figure*}[t]
\centering
\includegraphics[width=0.9\textwidth]{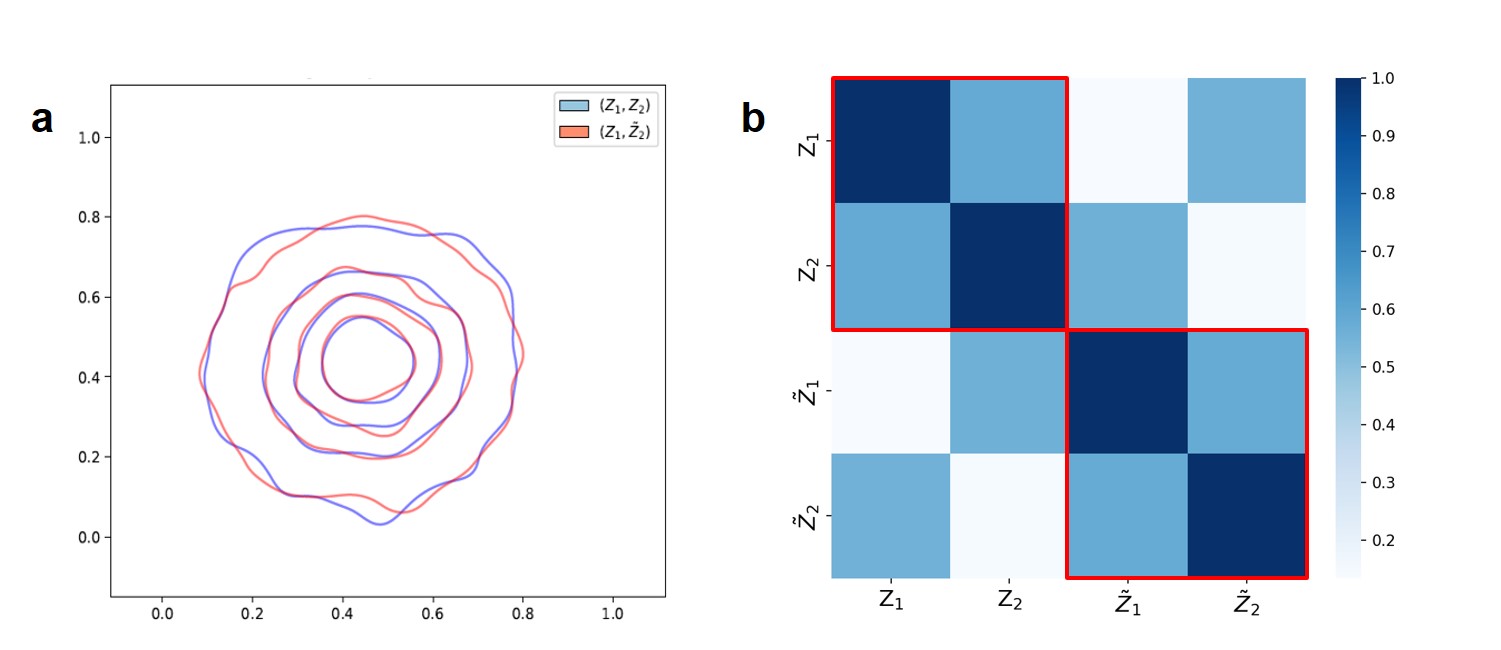}%
\caption{\textbf{a}. Illustration of exchangeability property $(Z_1, Z_2) \overset{d}{=} (Z_1, \tilde{Z}_2)$ of knockoff variables \textbf{b}. Shows the correlation matrix $\Sigma_{Z\tilde{Z}}$ of the original variables $Z$ and knockoffs $\tilde{Z}$ where the sub-matrix for the correlation within the generated knockoffs $\Sigma_{\tilde{Z}_{1}\tilde{Z}_{2}}$ is similar to that of correlation in original variables $\Sigma_{Z_{1}Z_{2}}$ , enclosed in red squares. While the variable-wise correlation is minimized with their respective knockoff copies, i.e., $\sigma_{Z_{1}\tilde{Z}_{1}}, \sigma_{Z_{2}\tilde{Z}_{2}}\approx 0$, shown in white squares.} 
\label{fig:knockoffs}
\end{figure*}

where $\tilde{X_i}$ is a knockoff representation or configuration to which the group of variables $X_i$ is set. Note that group $X_i$ contains internal process or nodes $Z_t, t=1, \dots, n$. Each node within a group is replaced with the generated knockoff copies. Applying the do-operator on trained deep networks involves specifying interventions or changes to the input group of time series to observe the corresponding effects on the target group. This implies that, instead of observing the natural variation in $X_i$, we are setting it to knockoff version $\tilde{X_i}$ in order to assess the collective causal impact on the network's output. We choose to use multivariate Gaussian model within the Knockoff framework  which implements semidefinite programming (SDP) to estimate knockoff parameters given the mean $\mu_Z$ and covariance matrix $\Sigma_Z$ of the original variables $Z$. We generate knockoff copies $\widetilde{Z}_1,\dots,\widetilde{Z}_{N}$ as an interventional representation of $Z_1, \dots, Z_{N}$, which are used to replace variables during an intervention. The generated knockoff variables are in-distribution and uncorrelated with the original variables. Moreover, it fulfills the exchangeability condition $(Z_i, \tilde{Z}_j) \overset{d}{=} (\tilde{Z}_i, Z_j), i,j=1,\dots, N, i\neq j$ \cite{barber2019knockoff} by maintaining same covariance structure as the original time series, which means that the underlying joint distribution of the data does not change by replacing a variable with its knockoff copy. The properties of knockoffs are demonstrated in Figure \ref{fig:knockoffs}.
\begin{table*}
\centering
\caption{Performance of the applied methods on synthetic data for varying interaction densities. The presented values are the ratio of the method output and the total number of tests where the \textit{Correct} inference represents the true causal direction, \textit{Wrong} is the opposite of the true causal direction, and \textit{No inference} is reporting the absence of a causal link.}
\begin{tabular}{lllllllllll}
    \hline
     & & \multicolumn{8}{c}{\textbf{Interaction density}}\\
     \cline{3-11}
     \textbf{Method} & \textbf{Inference} & 0.2 & 0.3 & 0.4 & 0.5 & 0.6 & 0.7 & 0.8 & 0.9 & 1.0  \\
    \hline
    \hline
    \multirow{2}{*}{gCDMI} & Correct & \textbf{1.00} & 0.67 & \textbf{1.00} & \textbf{1.00} & \textbf{1.00} & \textbf{0.67} & 0.67 & \textbf{1.00} & \textbf{1.00} \\
    \cline{2-11}
    & Wrong  & \textbf{0.00} & 0.33 & \textbf{0.00} & \textbf{0.00} & \textbf{0.00} & 0.33 & 0.33 & \textbf{0.00} & \textbf{0.00} \\
    \cline{2-11}
    & No inference  & \textbf{0.00} & \textbf{0.00} & \textbf{0.00} & \textbf{0.00} & \textbf{0.00} & \textbf{0.00} & \textbf{0.00} & \textbf{0.00} & \textbf{0.00}  \\
    \hline
    \hline
    \multirow{2}{*}{Trace} & Correct & 0.67 & \textbf{1.00} & 0.67 & 0.67 & 0.67 & \textbf{0.67} & 0.67 & \textbf{1.00} & 0.67  \\
    \cline{2-11}
    & Wrong &  0.33 & \textbf{0.00} & 0.33 & 0.33 & 0.33 & 0.33 & 0.33 & \textbf{0.00} &   0.33 \\
    \cline{2-11}
    & No inference  & \textbf{0.00} & \textbf{0.00} & \textbf{0.00} & \textbf{0.00} & \textbf{0.00} & \textbf{0.00} & \textbf{0.00} & \textbf{0.00} & \textbf{0.00}  \\
    \hline
    \hline
    \multirow{2}{*}{2GVecCI} & Correct & 0.67 & 0.00  & \textbf{1.00} &   0.33 & 0.67 & 0.33 & \textbf{1.00} & 0.67 & 0.00   \\
    \cline{2-11}
    & Wrong  & 0.33 & 0.33 & \textbf{0.00} &  0.33 & \textbf{0.00} & \textbf{0.00} & \textbf{0.00} & \textbf{0.00} & 0.33 \\
    \cline{2-11}
    & No inference  & \textbf{0.00} & 0.67 & \textbf{0.00} &  0.33 & 0.33 & 0.67 & \textbf{0.00} & 0.33 & 0.67 \\
    \hline
    \hline
     \multirow{2}{*}{Vanilla-PC} & Correct & 0.67 & 0.67 & 0.67 & 0.33 & 0.67 & \textbf{0.67} & 0.33 & 0.33 & 0.00   \\
    \cline{2-11}
    & Wrong  & \textbf{0.00} & \textbf{0.00}  & 0.33 & 0.33 & \textbf{0.00} & 0.33 & 0.67 & 0.67 & 1.00  \\
    \cline{2-11}
    & No inference  & 0.33 & 0.33 & \textbf{0.00} & 0.33 & 0.33 & \textbf{0.00} & \textbf{0.00} & \textbf{0.00} & \textbf{0.00} \\
    \hline
    
\end{tabular}
\label{tab:synresults}
\end{table*}
\paragraph{Group Causal Inference} In order to infer causal direction in two groups of variables, i.e., $X_i \rightarrow X_j$, we apply Kolmogorov–Smirnov (KS) test \cite{smirnov1939estimation} 
\begin{equation}
\label{eqn:kstest}
C_{ij}=\sqrt{\frac{qr}{q + r}}sup|R - \widetilde{R}|
\end{equation}
to evaluate model invariance property by estimating the maximum absolute difference in the marginal residual distribution $R$ and $\widetilde{R}$ of variables in $X_j$ before and after group-level intervention on $X_i$. Here $\widetilde{R}$ represents the distribution of the residuals $\widetilde{e}^n=\widetilde{e}_1, \dots, \widetilde{e}_n$, obtained while estimating the counterfactuals $P(Z_{i, t_0:T} | Z_{i, 1:t_0 - 1}, \widetilde{Z}_{-i, 1:t_0 - 1})$. The KS test uses a significance level $\alpha$ and other parameters like $q, r$ which represents the number of samples in marginal residual distributions $R$ and $\widetilde{R}$. The test statistic $C$ is used to calculate $p$ value for deciding whether a causal link exists or not. If $ p_{ij} > \alpha $, the null hypothesis $H_0$:  $X_i$ does not cause $X_j$ is accepted, which means that the residual distributions of group of variables before and after intervention are approximately identical across various interventional settings, i.e., the group invariance property is fulfilled. In the alternate case, the null hypothesis is rejected, i.e. $H_1$: $X_i$ causes $X_j$. We test for both causal links $X_{i} \rightarrow X_{j}$ and $X_{j} \rightarrow X_{i}$ for all possible values of $i, j$ where $i\neq j$. It is important to mention that if a single node within the group is influenced by its causal group, it establishes a causal connection between the groups and it doesn't necessarily imply that the entire target group is uniformly affected by its cause.
\begin{table}[t]
\centering
\small
\caption{Performance of different group causality methods in identifying causal direction in climate-ecosystem data for various sites. Here $\checkmark$ indicates the presence of the link, while $\times$ indicates the absence of the link and  $\rightarrow$ represents the direction of causal relation between groups. We test a bi-directional causal link only for our method.}
\begin{tabular}{llllll}
    \hline
     & & \multicolumn{4}{c}{\textbf{Methods}}\\
     \cline{3-6}
     \textbf{Sites} & \textbf{Inference} & \textbf{gCDMI} & \textbf{Trace} & \textbf{V-PC} & \textbf{2GVCI} \\
    \hline
    \hline
    \multirow{2}{*}{DE-Hai} & $G_{\text{C}} \rightarrow G_{\text{E}}$ & $\times$ & $\times$ & $\times$ &  $\times$ \\
    \cline{2-6}
    & $G_{\text{C}} \leftarrow G_{\text{E}}$  & $\times$ & $\checkmark$ & $\times$ & $\times$ \\
    \cline{2-6}
    & $G_{\text{C}} \leftrightarrow G_{\text{E}}$  & $\checkmark$ & $-$ & $-$ & $-$  \\
      \cline{2-6}
    & $G_{\text{C}} \nleftrightarrow G_{\text{E}}$  & $\times$ & $\times$ & $\checkmark$ & $\checkmark$  \\
    \hline
    \hline
    \multirow{2}{*}{IT-MBo} & $G_{\text{C}} \rightarrow G_{\text{E}}$ & $\checkmark$ & $\checkmark$ & $\checkmark$ &  $\times$  \\
    \cline{2-6}
    & $G_{\text{C}} \leftarrow G_{\text{E}}$  & $\times$ & $\times$ & $\times$  & $\times$ \\
    \cline{2-6}
    & $G_{\text{C}} \leftrightarrow G_{\text{E}}$  & $\times$ & $-$ & $-$ & $-$  \\
    \cline{2-6}
    & $G_{\text{C}} \nleftrightarrow G_{\text{E}}$  & $\times$ & $\times$ & $\times$ & $\checkmark$  \\
    \hline
    \hline
    \multirow{2}{*}{FR-Pue} & $G_{\text{C}} \rightarrow G_{\text{E}}$ & $\checkmark$ & $\times$ & $\times$ & $\times$  \\
    \cline{2-6}
    & $G_{\text{C}} \leftarrow G_{\text{E}}$  & $\times$ & $\checkmark$ & $\times$ & $\times$ \\
     \cline{2-6}
    & $G_{\text{C}} \leftrightarrow G_{\text{E}}$  & $\times$ & $-$ & $-$ & $-$  \\
    \cline{2-6}
    & $G_{\text{C}} \nleftrightarrow G_{\text{E}}$  & $\times$ & $\times$ & $\checkmark$ & $\checkmark$  \\
    \hline
    \hline
     \multirow{2}{*}{US-Ton} & $G_{\text{C}} \rightarrow G_{\text{E}}$ & $\checkmark$ & $\times$ & $\times$ & $\times$  \\
    \cline{2-6}
    & $G_{\text{C}} \leftarrow G_{\text{E}}$  & $\times$ & $\checkmark$ & $\times$ & $\checkmark$ \\
     \cline{2-6}
    & $G_{\text{C}} \leftrightarrow G_{\text{E}}$  & $\times$ & $-$ & $-$ & $-$  \\
    \cline{2-6}
    & $G_{\text{C}} \nleftrightarrow G_{\text{E}}$ & $\times$ & $\times$ & $\checkmark$ & $\times$ \\
    \hline
    
\end{tabular}
\label{tab:realdataresults}
\end{table}
\section{Experiments}
\label{experiments}
\paragraph{Synthetic Data} To evaluate the performance of our method, we use synthetic data model $Z_t^j = \Sigma_i  f_i (Z_{t-k}^i) + \eta_t^j$, $i, j = {1, ..., N}, 0<k<t$. The system variables $Z_j$ have auto and cross-functional dependencies with a time delay of  $k$. The data model incorporates linear and nonlinear dependencies $f$, i.e., exponential, polynomial with varying edge densities, and adds uncorrelated, normally distributed noise $\eta_t^j$. We incorporate inter- and intra-group causal links in the generated causal graphs. The decision of the methods for each edge in the groups is either \textit{correct}, \textit{wrong} or \textit{no inference}. Our method (gCDMI) achieves better identification of the correct causal links with fewer wrong detections in subsystems for all edge densities, i.e., sparse to dense causal graphs, as compared to other group causality methods as shown in Table \ref{tab:synresults}. The given values are the ratio of the method decision and the total number of experiments. It can be noticed that our method reported only fewer wrong inferences and zero \textit{no inference} along with the Trace method for all experiments, while other methods reported \textit{no inference} in a number of tests because they are uni-directional and sometimes it is hard to distinguish cause from effect when the interaction in both directions is not significantly different. We also found that increasing edge density does not have a clear impact on the outcome except for Vanilla-PC, where the performance is degraded. We used multiple group dimensions for each edge density in our experiments, where we vary the architecture of the deep networks based on group dimension. All experiments in this paper were conducted on an NVIDIA GeForce RTX 3060 Ti graphics card, providing reliable performance capabilities for complex synthetic and real-world data causal analysis. The improved performance of our method comes at the cost of high computation time due to its dependence on deep networks to learn complex causal structures in data.

\paragraph{FLUXNET Data} 
Here, we aim to perform a causal analysis of environmental time series with our method by establishing the causal structure in groups or subsystems. We carry out experiments with FLUXNET2015 dataset \cite{pastorello2020fluxnet2015}, which is acquired using the eddy covariance technique to measure the cycling of carbon, water, and energy between the biosphere and atmosphere through collaborations among many regional networks, with data preparation efforts happening at site and network levels. For our experiments, we considered various measurement sites, i.e., Hainich (DE-Hai: Deciduous Broadleaf Forests), Monte Bondone (IT-MBo: Grasslands), Puechabon (FR-Pue: Evergreen Broadleaf Forests), and Tonzi Ranch (US-Ton: Woody Savannas) site. The dataset includes climatic and ecological time series, i.e., global radiation ($R_\text{g}$), temperature ($T$), gross primary production ($GPP$), ecosystem respiration ($R_{\text{eco}}$) for various time scales i.e., half-hourly, hourly, daily, weekly and so on. We categorized these variables into two groups: climate group  $G_{\text{C}}$ which contains $T$ and $R_g$ and ecosystem group $G_{\text{E}}$, which consists of the ecosystem variables $GPP$ and $R_{\text{eco}}$. We considered daily sampling which is advantageous for mitigating the effects of daily patterns that usually undermine the underlying causal relation in data. The results for all applied causality methods from various sites are given in Table \ref{tab:realdataresults}. Our method identified the presence of a causal link $G_{\text{C}} \rightarrow G_{\text{E}}$ for all sites except for the DE-Hai site where we obtain a bidirectional link $G_{\text{C}} \leftrightarrow G_{\text{E}}$ which is an overall better performance compared to other methods. The bidirectional link at the DE-Hai site, which is a deciduous broadleaf forest, could be due to climate-ecosystem strong feedback mechanism or other influential factors that need further investigation. For DE-Hai, Vanilla-PC and 2GVecCI could not infer any causal direction while the Trace method detected $G_{\text{E}} \rightarrow G_{\text{C}}$. For IT-MBo, all methods correctly identified the expected causal directions except 2GVecCI. As an illustration, we show climatic influence on the ecosystem for the IT-MBo site from one of our experiments in Figure \ref{fig:jointdistshift} (a). 

\paragraph{ENSO Data} Moreover, we performed experiments on the ENSO dataset where we consider surface temperatures over the ENSO region and British Columbia (BCT) from 1948 to 2021, as a causal effect of temperatures in the tropical Pacific on those in North America is recognized in climate \cite{taylor1998effect}. During an El Niño event, which is one phase of ENSO, the tropical Pacific Ocean warms up significantly, disrupt normal weather patterns and influence the climate in other parts of the world, including the British Columbia region. For experiments, we adapted data preprocessing, i.e., deseasonalizing, smoothing, and aggregation from the work of \cite{runge2019detecting, wahl2023vector}. We show results for applied group causality methods in Table \ref{tab:enso}. Both our method and 2GVecCI identified significant influence of ENSO on BCT for various grid scales with a fraction of 0.66 correct inferences and 0.34 as no inferences. While Trace method detected the causal link ENSO $\rightarrow$ BCT with a fraction of 0.50 correct inferences and 0.50 wrong inferences. Vanilla-PC could not inferred any causal direction for all grid scales that is probably because of difficultly in distinguishing data patterns at both regions. Illustration of the causal influence of ENSO on BCT for one of the performed tests by our method is given in Figure \ref{fig:jointdistshift} (b). Which shows the comparison of the BCT distribution with and without group-level intervention on the ENSO time series variables. The shift in the counterfactual distribution for BCT is indicative of the influence of ENSO.

\begin{table}[hbt!]
\centering
\caption{Effect of surface temperature at Tropical Pacific ocean (ENSO region) on British Columbia region, computed at various grid scales. The given values represent the decision of the applied methods (in percentage).}
\begin{tabular}{lllll}
    \hline
    \textbf{Inference} & \textbf{gCDMI} & \textbf{Trace} & \textbf{2GVCI} & \textbf{V-PC} \\
    \hline
    ENSO $\rightarrow$ BCT & \textbf{0.66} & 0.50 & \textbf{0.66} & 0.00\\
    \hline
    ENSO $\leftarrow$ BCT & \textbf{0.00} & 0.50 & \textbf{0.00} & \textbf{0.00}\\
    \hline
    ENSO $\nleftrightarrow$ BCT & 0.34 & \textbf{0.00} & 0.34 & 1.00\\
    \hline
    
\end{tabular}
\label{tab:enso}
\end{table}
%

\paragraph{fMRI Data} We analyzed simulated fMRI time series data \cite{smith2011network} to identify connections in brain network nodes and assign direction to them. These fMRI time series simulations were based on dynamic causal modelling (DCM) \cite{friston2003dynamic} fMRI forward model. The generated dataset provides ground truth connectivity graph for variety of network topologies which we use to evaluate our methods. Here we conducted experiments on a number of simulated subjects from \textit{S}2 topology in the dataset which contains 10 nodes clustered into 2 groups. It has 10 min fMRI sessions for each subject with 3 sec sampling rate, ﬁnal added noise of 1\%, and haemodynamic response function (HRF) variability of ±0.5 which provides time series of 200 data points. The percentage of methods outcome in terms of \textit{correct}, \textit{wrong} and \textit{no inference} is given in Table \ref{tab:fmri}. Since our method is data demanding as it relies on deep networks, the provided size of time series was not long enough for our method to learn the complex relationship properly. Determining the correct directionality of connections in fMRI time series is challenging because of the complex interactions among brain networks. However, our approach inferred correct direction 56\% of the times in brain networks with 21\% bidirectional links which is overall better performance compared to other applied group causality methods. Results from Trace method were somehow comparable to that of gCDMI. While 2GVecCI and Vanilla-PC yielded high percentage of \textit{no inference} probably because of difficulty in separating cause from effect in a mutual interaction scenario between groups. The causal influence of brain network $N_1$ on network $N_2$ for one subject in \textit{S}2 topology from simulated fMRI time series is demonstrated in Figure \ref{fig:jointdistshift} (c).
\begin{table}[hbt!]
\centering
\caption{Identification of the direction of brain network connections using simulated fMRI time series by the applied group causality methods: Ground truth network connection: $N_1 \rightarrow N_2$. The given values represent the decision of the applied methods (in percentage).}
\begin{tabular}{lllll}
    \hline
    \textbf{Inference} & \textbf{gCDMI} & \textbf{Trace} & \textbf{2GVCI} & \textbf{Vanilla-PC} \\
    \hline
    $N_1 \rightarrow N_2$ & \textbf{0.56} & 0.50 & 0.33 & 0.06\\
    \hline
    $N_1 \leftarrow N_2$ & 0.17 & 0.39 & 0.17 & \textbf{0.16}\\
    \hline
    $N_1 \leftrightarrow N_2$ & \textbf{0.21} & - & - & - \\
    \hline
    $N_1 \nleftrightarrow N_2$ & \textbf{0.06} & 0.11 & 0.50 & 0.78\\
    \hline
    
\end{tabular}
\label{tab:fmri}
\end{table}
%


\section{Conclusion}
\label{conclusion}

In this work, we have introduced a deep learning-based method to uncover the causal interactions in groups or subsets of time series. We model complex relationships in an $N$-variable system of time series using deep learning models, and apply invariance testing via group-level intervention to infer causal direction in groups of variables. Our approach also tests for bi-directional causal links which signify a mutual influence, suggesting that the variables have a reciprocal cause-and-effect relationship. While our approach demonstrated improved performance on both synthetic and real-word time series data compared to other causality methods, it is worth noting that the high computational time is a trade-off inherent to its reliance on deep learning. As future work, we consider estimating the causal interactions in more than two groups of time series. Moreover, we aim to address the issue of non-stationarity and hidden confounding \cite{trifunov2022time} to further improve our method.
\section{Acknowledgments}

This work is funded by the German Research Foundation (DFG) research grant SH 1682/1-1 and the Carl Zeiss Foundation within the scope of the program line \say{Breakthroughs: Exploring Intelligent Systems} for \say{Digitization — explore the basics (No P2017-01-003), use applications}. 



\end{document}